\documentclass[letterpaper, 10 pt, conference]{ieeeconf}  

\IEEEoverridecommandlockouts                              

\usepackage{times}
\usepackage{epsfig}
\usepackage{graphicx}
\usepackage{graphics} 
\usepackage{float}
\usepackage{epsfig} 
\usepackage{times} 
\usepackage{amsmath} 
\usepackage{amssymb}  
\usepackage{booktabs}
\usepackage{epstopdf}
\usepackage{color}
\usepackage{url}
\usepackage{cite}
\usepackage{balance}
\usepackage{textcomp} 
\usepackage{caption}
\usepackage{makecell}

\title{\LARGE \bf
A Multi-spectral Dataset for Evaluating Motion Estimation Systems
}

\author{Weichen~Dai$^{1}$,~Yu~Zhang$^{1}$,~Shenzhou Chen$^{2}$,~Donglei~Sun$^{3}$,~and~Da~Kong$^{1}$~
\thanks{$^{1}$State~Key~Laboratory~of~Industrial~Control~Technology, College~of~Control~Science~and~Engineering, Zhejiang University, 
{\tt \scriptsize \{weichendai,zhangyu80,kongda1997\}@zju.edu.cn}}%
\thanks{$^{2}$Alibaba A.I. Labs, {\tt \scriptsize ~shenzhou.csz@alibaba-inc.com}}
\thanks{$^{3}$Centre for English Language Education, University~of~Nottingham Ningbo China, 
{\tt \scriptsize donglei.sun@nottingham.edu.cn}}
\thanks{This work was supported by NSFC 62088101 Autonomous Intelligent Unmanned Systems, the National Natural Science Foundation of China (Grant No. 61673341), National Key R\&D Program of China (2016YFD0200701-3), Double First Class University Plan (CN), the Project of State Key Laboratory of Industrial Control Technology, Zhejiang University, China (No.ICT2021A10).}
}

\begin{document}

\maketitle
\thispagestyle{empty}
\pagestyle{empty}

\begin{abstract}

Visible images have been widely used for motion estimation. Thermal images, in contrast, are more challenging to be used in motion estimation since they typically have lower resolution, less texture, and more noise.
In this paper, a novel dataset for evaluating the performance of multi-spectral motion estimation systems is presented.
All the sequences are recorded from a handheld multi-spectral device. It consists of a standard visible-light camera, a long-wave infrared camera, an RGB-D camera, and an inertial measurement unit (IMU).
The multi-spectral images, including both color and thermal images in full sensor resolution (640 $\times$ 480), are obtained from a standard and a long-wave infrared camera at 32Hz with hardware-synchronization.
The depth images are captured by a Microsoft Kinect2 and can have benefits for learning cross-modalities stereo matching.
For trajectory evaluation, accurate ground-truth camera poses obtained from a motion capture system are provided.
In addition to the sequences with bright illumination, the dataset also contains dim, varying, and complex illumination scenes.
The full dataset, including raw data and calibration data with detailed data format specifications, is publicly available.

\end{abstract}

\section{INTRODUCTION}

In recent years, vision-based motion estimation methods such as visual odometry (VO) \cite{scaramuzza2011visual} and visual simultaneous localization and mapping (vSLAM) \cite{fuentes2015visual} have attracted full attention for their diverse applications. 
These methods have been investigated in great detail for standard cameras, which can take advantage of rich textures only in bright illumination. For example, in scenarios such as data center inspection, firefighting, and rescue, standard cameras cannot provide sufficient information due to inadequate color textures, smog cover, or dim illumination. 
Therefore, some different types of sensors are used to enhance the robustness of vision-based motion estimation methods.
In environments with complex illumination conditions, adding long-wave infrared (LWIR) cameras can complement the texture with information from another spectrum. Hence, the multi-spectral setup with these two types of cameras can become a reliable information source for all-day vision or fog-penetrating localization.

To evaluate the performance of various multi-spectral SLAM and odometry methods involving multi-spectral sources, a complete dataset with ground truth is necessary. Compared with the dataset for stereo standard cameras \cite{sturm2012benchmark, geiger2013vision, schubert2018tum}, the availability of a hardware-synchronization multi-spectral dataset is minimal. Moreover, stereo matching between RGB images and thermal images remains a challenging problem due to the insufficiency of textures in the latter and the difference between modalities. Therefore, the dataset should also provide reference stereo correspondences between different modalities.

In this paper, a multi-spectral dataset is provided for evaluating multi-spectral motion estimation methods. The dataset includes a set of sequences in diverse illumination conditions obtained on a setup consisting of a standard camera, an LWIR camera, an IMU, and a Kinect2. In the dataset, each sequence contains the color, thermal, and depth images, as well as the ground-truth trajectory. The color and thermal images (640 $\times$ 480) are captured at 32Hz with hardware-synchronization. The dense depth images are provided by the Kinect2 camera at 30Hz, enabling the projection of one image onto the other. The acceleration and angular velocity are measured by an Xsens IMU for challenging sequences. The ground-truth poses are recorded from a motion capture system at 120Hz. All sensors are calibrated carefully for higher accuracy.

The whole dataset, including the raw and calibration data, calibration files, and tool codes, is available on
\begin{center} {\color{red}\textit{https://github.com/NGCLAB/multi-spectral-dataset}}. \end{center}

The main contributions of this paper are as follows:
\begin{itemize}
	\item {A new hardware-synchronization multi-spectral dataset with groundtruth poses is provided for the evaluation of multi-spectral motion estimation systems.}
	\item The additional depth images can be used to study stereo matching between the visible and LWIR spectra.
\end{itemize}

The rest of the paper is organized as follows. Related work is reviewed in Section \ref{sec:related_work}, and the platform is introduced in Section \ref{sec:platform}. 
The details of time synchronization and evaluation are presented in Section \ref{sec:time_sync} and the dataset is described in \ref{sec:dataset}.
Finally, the experiments conducted in this dataset are discussed in Section \ref{sec:exp} and conclusions are drawn in Section \ref{sec:conclusion}.

\section{RELATED WORK}
\label{sec:related_work}
Several types of visual sensors have been used in motion estimation, including monocular standard cameras \cite{engel2017direct, forster2017svo}, stereo standard cameras \cite{campos2020orb}, event-cameras \cite{vidal2018ultimate}, and RGB-D cameras \cite{endres20133}. These motion estimation methods can be categorized into filter-based \cite{davison2007monoslam} and factor-graph-optimization-based methods \cite{strasdat2012visual}. Besides, methods~\cite{wang2017deepvo, zhou2017unsupervised} with deep learning methods have attracted the interest of the community in recent years. The majority of these methods focus on utilizing the information on the visible spectrum.

Since the performance of standard cameras can be significantly influenced by illumination, LWIR cameras with multiple configurations, such as omnidirectional thermal cameras~\cite{benli2019thermal} and stereo thermal cameras~\cite{mita2019robust}, have been explored. With a different sensor modality, LWIR cameras are illumination independent, but they have their shortcomings. As mentioned in \cite{hajebi2008structure}, the notable shortcomings include high noise level, high dynamic range, and the unique non-uniformity correction (NUC) mechanism that causes image corruption.
Due to these factors, the visual odometry using LWIR setups \cite{borges2016practical, mouats2015thermal} usually cannot provide accurate results as those methods based on visible light in most environments.
Hence, the methods~\cite{khattak2019keyframe, papachristos2018thermal, poujol2016visible} that integrate the information from IMU and those~\cite{zhao2020tp, saputra2020deeptio} that rely on deep learning algorithms present a more practical solution.
Besides, multi-spectral methods using LWIR cameras as complementary sensors have attracted extensive attention \cite{mouats2015multispectral,dai2019multi} for their potential to function in poorly illuminated environments.

For visible light sensors, there are several well-known datasets, where the illumination condition is usually stable. Depending on the application, the carriers may be handheld \cite{caruso2015_omni_lsdslam}, cars \cite{geiger2013vision}, micro aerial vehicles \cite{burri2016euroc}, and underwater vehicles \cite{ferrera2019aqualoc}. For the task of fusing multi-sensor information, most datasets provide synchronized sensor data in addition to visible images \cite{majdik2017zurich}. Meanwhile, since hardware synchronization of sensors is critical~\cite{leutenegger2015keyframe}, some high quality datasets also designed hardware-synchronization devices. Moreover, motion capture systems such as VICON are used to obtain the ground truth in most datasets. In other datasets, the ground-truth trajectories are acquired through GPS or accurate 3D reconstruction.

\begin{table*}[t!]
    \centering
    \caption{Comparison of multi-spectral datasets for motion estimation systems}
    \begin{tabular}{cccccccc}
    \toprule
         Dataset & Year & Environment & Carrier & Spectrum & Baseline & Time
         sync & Ground truth \\
         \midrule
         KAIST Odometry & 2018 & Urban & car & \thead{ RGB @25Hz\\ LWIR @25Hz }&  \thead {0.00m \\ parallax-free} & hw & \thead{ OXTS RT 2002 \\ pose @100Hz, acc. $<$2cm }\\
         Agricultural robot & 2017 & Terrain & car & \thead{ RGB @30Hz \\ NIR @30Hz }&  \thead {0.00m \\ parallax-free} & sw/hw & \thead{ Leica RTK/Ublox EVK7-PPP \\ position @10/4Hz, acc. $<$3/250cm }\\
         Vivid & 2019 & In-/outdoor & handheld & \thead{ RGB @30Hz \\ LWIR @20Hz}&  \thead {0.05m} & sw & \thead{ Cortex motion capture system\\/LeGO-LOAM \\ pose @100Hz, acc. $<$1cm }\\
         \textbf{Ours} & 2020 & \textbf{In-/outdoor} & handheld & \thead{ RGB @32Hz\\ LWIR @32Hz   }&  \thead {\textbf{0.14m}} & \textbf{hw} & \thead{ Optitrack motion capture system \\ pose @120Hz, acc. $<$1cm }\\
         \bottomrule
    \end{tabular}
    \label{tab:dataset_overview}
    \vspace{-16pt}
\end{table*}

Most of the existing multi-spectral datasets were generated for fundamental tasks such as detection \cite{hwang2015multispectral}, segmentation \cite{shivakumar2019pst900}, tracking \cite{li2019rgb}, stereo matching \cite{treible2017cats}, and place recognition \cite{maddern2012towards}. A publicly available dataset with ground truth is also critical for evaluating the performance of multi-spectral motion estimation methods. For the task of motion estimation, the KAIST multi-spectral dataset \cite{choi2018kaist} was proposed for all-day vision tasks in outdoor environments covering a wide range from urban to residential regions. In addition to RGB information, agricultural robots \cite{chebrolu2017ijrr}, also measure light emissions in the near-infrared (NIR) spectrum for separating vegetation from the soil and other background data. Both datasets are designed for outdoor environments.
However, in most indoor environments, since thermal images cannot provide rich textures, it is more challenging to design multi-spectral motion estimation. Although the Vivid dataset \cite{leevivid} provides multi-spectral images, it cannot provide data with hardware synchronization. 
As summarised in Table~\ref{tab:dataset_overview}, there is a lack of multi-spectral dataset with hardware synchronization for both indoor and outdoor environments.
Moreover, since stereo matching is the foundation of stereo methods, providing depth data plays an essential role in augmenting the development and validation of cross-modality matching algorithms.

\section{PLATFORM}
\label{sec:platform}

\begin{figure}[t]
	\centering
	\includegraphics[scale=0.8]{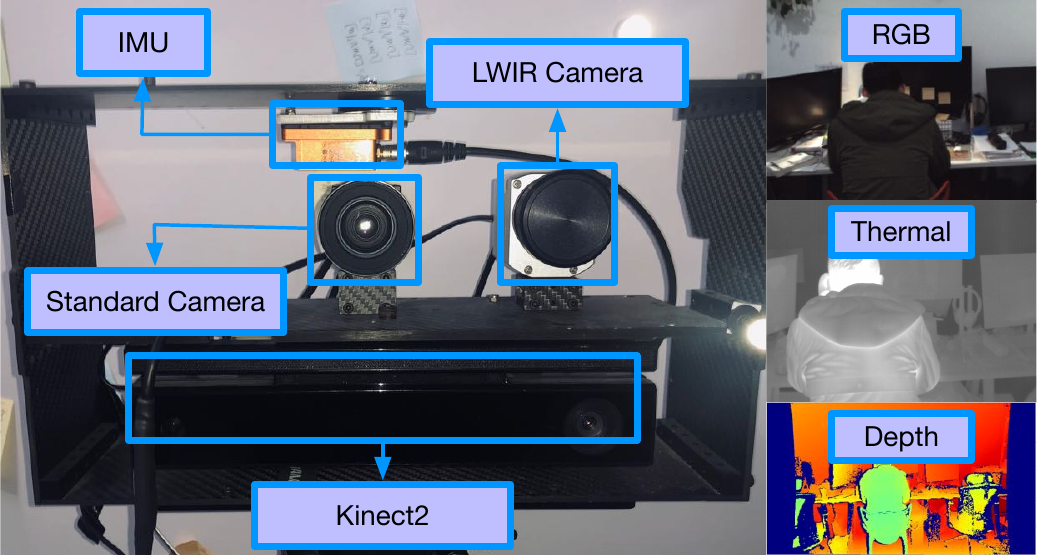}
	\caption{The multi-spectral setup with Kinect2. \textit{Left}: devices include an RGB camera, an LWIR camera, a Kinect2, an IMU, and the markers of motion capture systems. \textit{Right}: example color image, thermal image, and depth image from Kinect2.}
	\label{fig:setup}
	\vspace{-16pt}
\end{figure}

The multi-spectral setup consists of a standard camera, an LWIR camera, an IMU, and a Kinect2, as shown in Fig. \ref{fig:setup}. To record accurate ground truth for the dataset, the OptiTrack motion capture system is used to provide 6D poses of the multi-spectral device using the markers on the setup. 

The standard camera, LWIR camera, IMU, and Kinect2 are rigidly connected to a rack with OptiTrack markers. 
In addition to a printed chessboard, a unique chessboard whose surface is made of different materials is used to calibrate the system. Meanwhile, the parameters between the camera and the IMU are calibrated using kalibr~\cite{rehder2016extending} based on  a grid of AprilTags~\cite{olson2010apriltag}, which is static in the scene. 
The necessary sensor information is summarized in Table \ref{tab:setup}. In the following part, the key hardware component and how to calibrate it are briefly described.

\begin{table}[t]
	
	\centering
	\caption{Overview of sensors in the setup
	}
	\label{tab:setup}
	\resizebox{8.8cm}{!}{
	\begin{tabular}{c|ccc}
			\toprule
			Sensor &  Model  & Rate  &  Characteristics   \\ 
			\midrule
			\thead{Standard color camera}   & ImageSource   & 32Hz      & \thead{ global shutter  \\
			640$\times$480\\
		     RGB32} \\ \hline
			LWIR camera   & Optris 	& 32Hz         & \thead{ 130 mK \\
			640$\times$480\\
			16-bit thermal\\
			Spectral range 7.5-13 $\mu m$\\
			Non-uniformity correction }\\\hline
			Depth camera   & Kinect2 	& 30Hz    & \thead{Time of Flight  \\
			960$\times$540\\
			16-bit depth}\\ \hline
			IMU   & Xsens 	& 400Hz    & \thead{3D accelerometer   \\
			3D gyroscope}\\ \hline
			Motion caption system & OptiTrack   & 120Hz     & \thead{6-D pose  \\
			Four cameras} \\
			\bottomrule
	\end{tabular}}
	\vspace{-8pt}
\end{table}

\subsection{The multi-spectral device}

The multi-spectral device consists of a standard camera and an LWIR camera.
The former is an ImageSource DFK 22BUC03. It uses a global shutter and captures 640$\times$480 RGB images at 32 Hz. Moreover, this camera can be synchronized by an external trigger signal. The LWIR camera is an Optris PI 640, which produces 16-bit 640$\times$480 thermal images and outputs a frame-sync trigger signal at 32Hz. This frame-sync trigger signal is set as the external trigger signal via a hardware connection to the standard camera. Therefore, this platform can provide synchronized color and thermal images at 32Hz when both cameras are set on trigger mode. The exposure time is set to the value less than the sensor synchronization period, which ensures that the captured images are at the same frequency.

Due to the difference in information sources, the conventional printed board used to calibrate standard cameras cannot yield high contrast between textures in the thermal image. Therefore, a specialized equipment needs to be designed for calibration. There are two calibration methods: active \cite{zoetgnande2019robust, vidas2012mask} and passive \cite{mouats2015thermal}. Active methods heat part of the calibration equipment to generate different thermal radiations. Passive methods do not depend on active heating or cooling. 

\begin{figure}[t]
	\centering
	\begin{minipage}{0.23\textwidth}
	    \centering
		\includegraphics[width=\textwidth]  {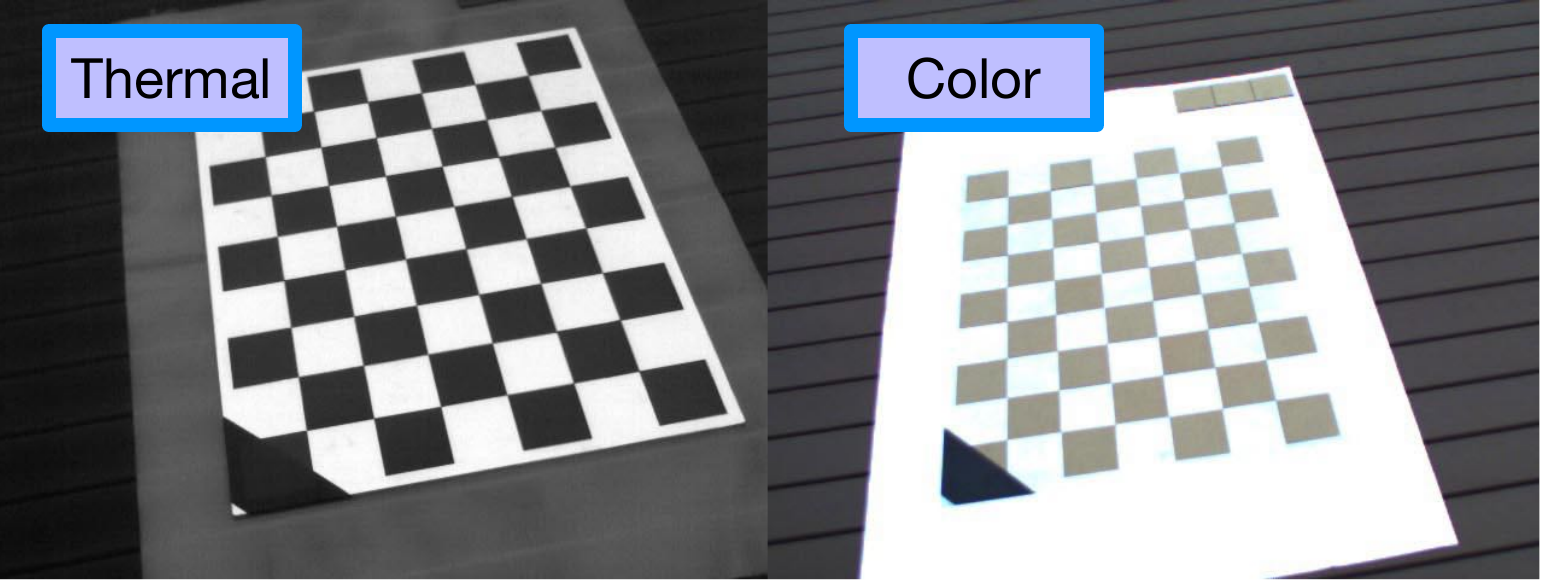}
		{\footnotesize(a) Our board}
	\end{minipage}
	\begin{minipage}{0.23\textwidth}
	    \centering
		\includegraphics[width=\textwidth]  {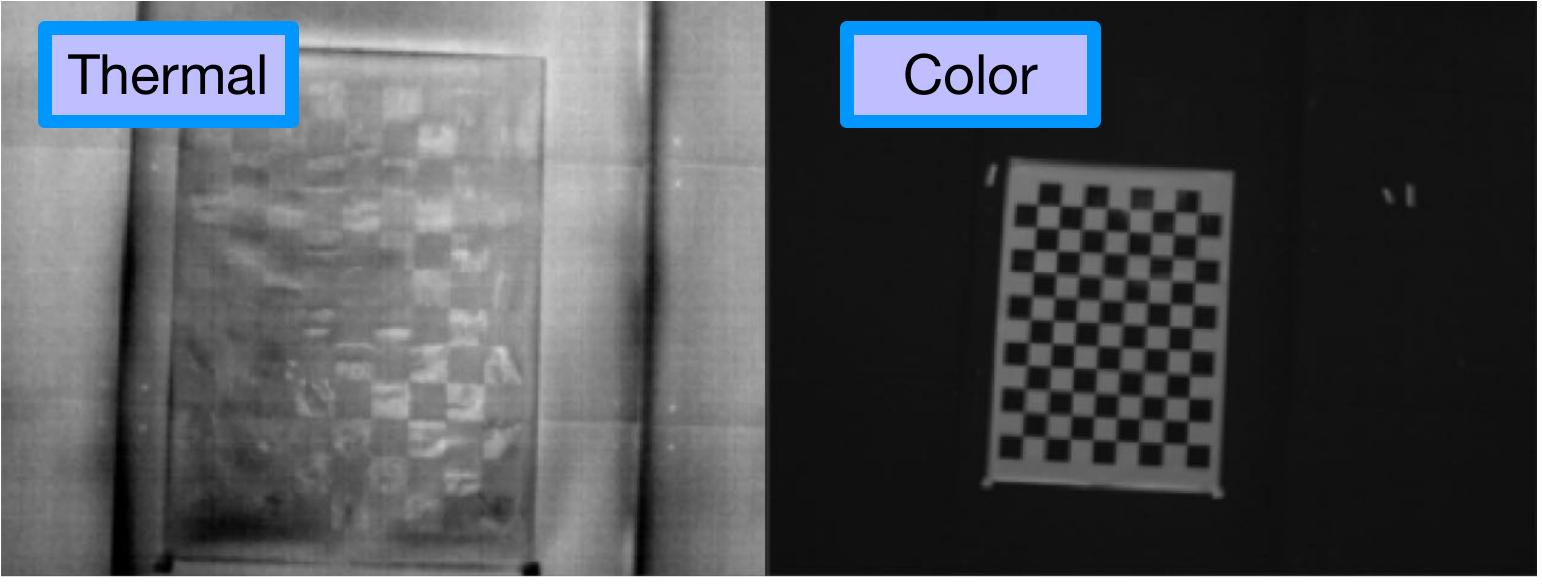}
		{\footnotesize (b) Printed board}
	\end{minipage}
	\caption{Comparison of two calibration boards. A unique chessboard with the surface made of different materials can provide distinctive textures in both spectra. In contrast, the chessboard on the printed board cannot create distinct grids in the thermal camera.}
	\label{fig:ms_calib_board}
	\vspace{-16pt}
\end{figure}

Since active methods are inconvenient and complicated, a passive method is used for this setup. Exploiting the difference in the reflectivity of metallic and non-metallic materials in both the LWIR and visible spectra, an aluminum board with fiber squares on it is designed for calibration. This board provides distinct edges and excellent contrast in the sun, as shown in Fig. \ref{fig:ms_calib_board}(a). During the calibration, the device followed a stop-capture-go manner to eliminate the error introduced by unsynchronized data. The intrinsic and extrinsic parameters of the multi-spectral device are obtained using MATLAB\textsuperscript{\textregistered} tools~\cite{zhang2000flexible}. 

\subsection{Kinect2}

To provide a reference for stereo matching between the visible and LWIR spectra, a Kinect2 is added to the setup. It integrates both the color and the depth camera. The depth images are 960$\times$540, captured at 30Hz. Since Kinect2 follows the time of flight (ToF) measurement principle, the phase difference between the infrared illuminator and the infrared camera is used to compute the depth. The timestamp of the depth image is well aligned with the multi-spectral device by the software.

The infrared camera of Kinect2 is a near-infrared (NIR) camera. Since the NIR spectrum is next to the visible spectrum, both spectra share similar textures. Therefore, the printed chessboard can be directly used to calibrate these two cameras. The difficulty in image capture lies in that the NIR light emitter cannot be turned off, and hence additional effort is required. Meanwhile, to obtain clear NIR texture, the illumination should also be adjusted appropriately.

\begin{figure}[t]
	\centering
	\begin{minipage}{0.23\textwidth}
	    \centering
		\includegraphics[width=\textwidth]  {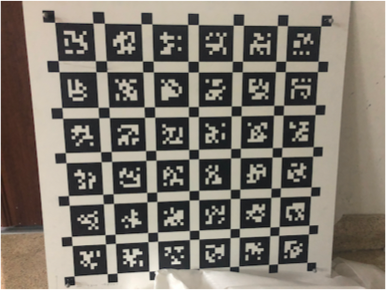}
		{\footnotesize (a) AprilTags  board }
	\end{minipage}
	\begin{minipage}{0.23\textwidth}
	    \centering
		\includegraphics[width=\textwidth]  {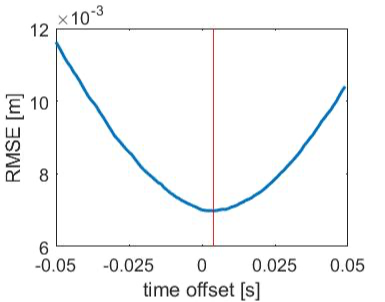}
		{\footnotesize (b) Analysis of time delay}
	\end{minipage}
	\caption{(a) AprilTags board used for calibration and time synchronization. Four reflective markers are attached to the four small corner squares of the board. (b) The time delay between the motion capture system and the standard camera.}
	\label{fig:aprilTags}
	\vspace{-16pt}
\end{figure}

\section{TIME SYNCHRONIZATION}
\label{sec:time_sync}

In the data sequences, every sensor should be synchronized with the standard camera, since incorrectly paired images will introduce error. 

\begin{figure}[t]
	\centering
	\includegraphics[scale=0.35]{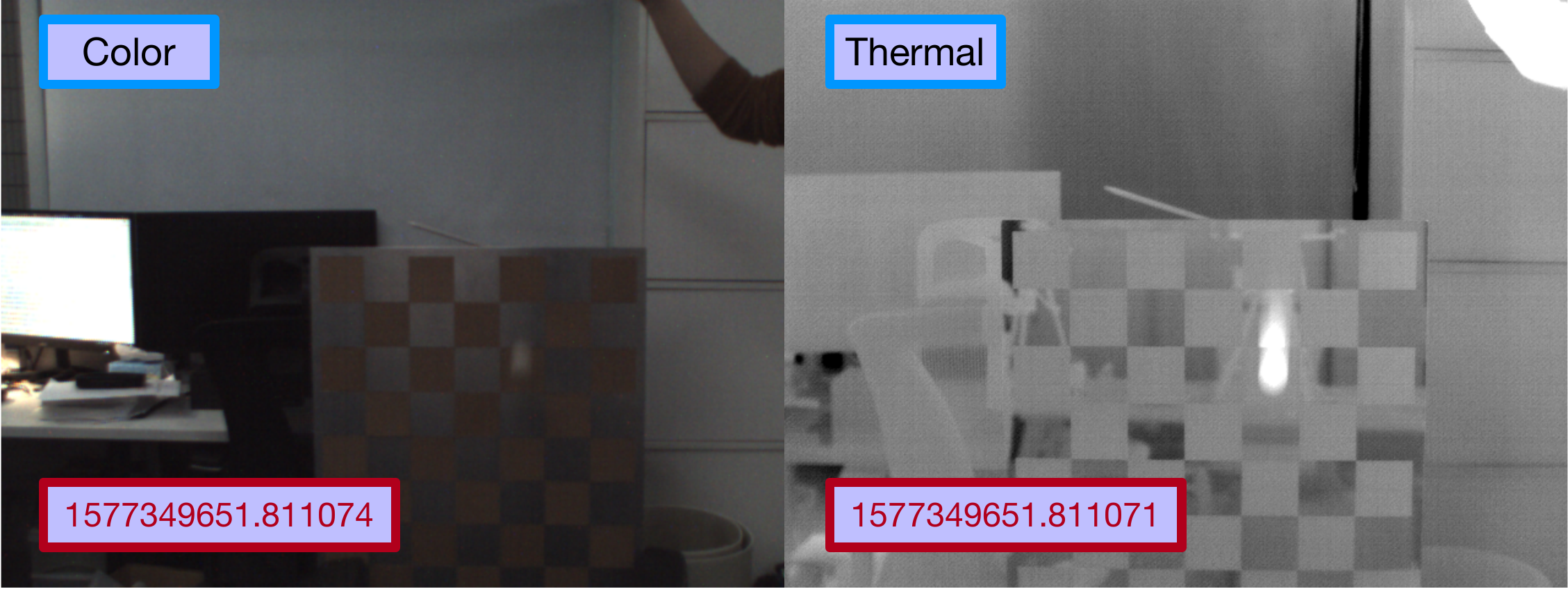}
	\caption{Time-synchronization evaluation of multi-spectral images. In both images with the same timestamp, the position of the dropping ball indicates that two cameras can capture images almost simultaneously. Among most pairs, the thermal images are captured approximately 1ms later than the color images. }
	\label{fig:time_sync_test}
	\vspace{-8pt}
\end{figure}

High-accuracy hardware-synchronization is achieved on the multi-spectral device using the frame-sync signal generated by the LWIR camera. Upon the rising pulse, the image from the standard camera is captured. An experiment was designed to check the performance of the synchronization. In this experiment, a ball was released in front of the board. As shown in Fig. \ref{fig:time_sync_test}, both cameras can capture the same scene with little time delay. The actual difference between the timestamps of those two images is less than 1ms. Hence, no modifications to the timestamps are required in the dataset. 


\begin{figure}[t]
	\centering
	\includegraphics[scale=0.35]{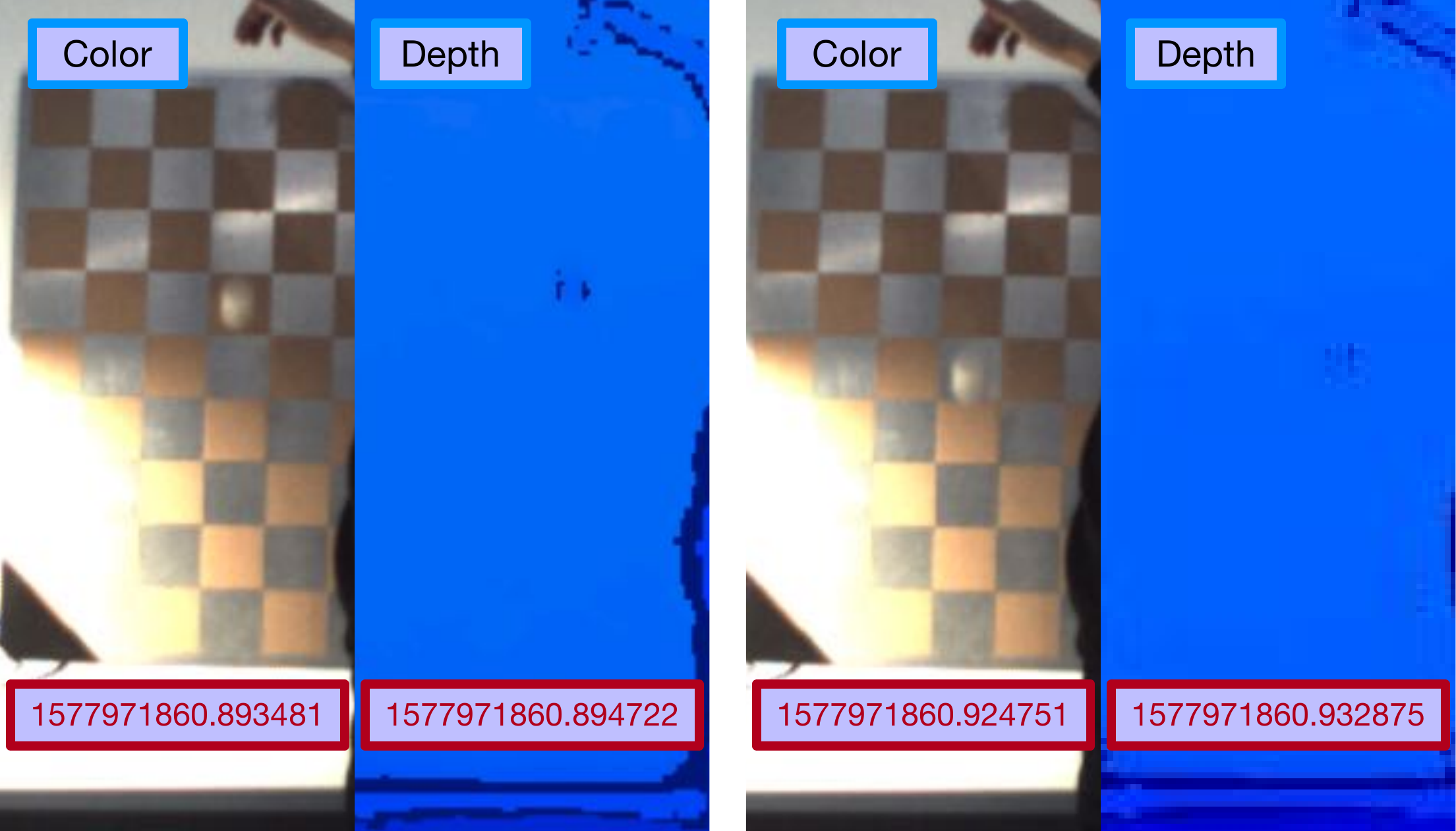}
	\caption{Time-synchronization evaluation of Kinect2. Images from the standard camera and the depth camera. The timestamps in those red boxes indicate that the depth images are captured slightly later than the color images. }
	\label{fig:kinect_sync_test}
	\vspace{-16pt}
\end{figure}

For the Kinect2 and the standard camera, the depth images are captured on average a little later than the color images, as shown in Fig.~\ref{fig:kinect_sync_test}. The delay in the depth images is negligible, and no correction is required before data association.


The time delay between the motion capture system and the standard camera also needs to be determined before localization results can be evaluated. The time delay can be determined from the residuals of different time delays. As shown in Fig. \ref{fig:aprilTags}(b), the poses from the motion capture system are approximately 4ms earlier than the images from the standard camera. The time delay is also trivial, and hence it is not necessary to modify the timestamps of the raw poses.

\section{THE DATASET}
\label{sec:dataset}

The dataset includes both the evaluation sequences and the calibration data with a size of approximately 700GB. In the calibration data, both the raw data and our calibration results are provided.
The dataset includes the following categories:
\begin{itemize}
    \item Calibration:
    \begin{itemize}
	\item \textit{visible-LWIR}: the calibration data to compute the intrinsics and extrinsics of the multi-spectral device. The board 
	shown in Fig. \ref{fig:ms_calib_board} is recorded in a stop-capture-go manner with changing viewpoints and small camera motion.
	\item \textit{visible-Kinect}: the calibration data to find the intrinsics and extrinsics of Kinect2 and to obtain the transformations between Kinect2 and the standard camera.
	\end{itemize}
	\item Evaluation sequences: The sequences can be divided into two types.
	\begin{itemize}
	\item \textit{Testing and debugging scenes}: sequences intended to facilitate the development of novel multi-spectral algorithms with separated criteria, including motion, illumination, and person. For convenience, these sequences are divided into three categories according to the illumination:
	\begin{itemize}
	    \item \textit{bright scenes}: captured in a bright room where the color images contain rich textures.
	    \item \textit{scenes with varying illumination}: captured in a room with varying illumination, where the quality of the color images is not guaranteed.
	    \item \textit{dark scenes}: captured in a dark room. The color images do not provide clear visible textures. 
	\end{itemize}
	\item \textit{challenging scenes}: sequences captured from more challenging environments, including texture-less, high-dynamic-range, and dark areas. Therefore, these sequences are very challenging for multi-spectral methods. To reduce the difficulty, the IMU information is provided. 
    \end{itemize}
\end{itemize}

\subsection{Sequences}

\begin{figure}[t!]
	\centering
	\begin{minipage}{0.15\textwidth}
	    \centering
		\includegraphics[width=\textwidth]  {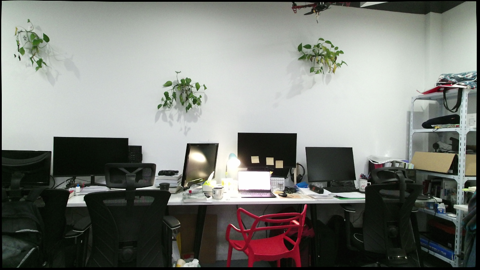}
		{\footnotesize (a) Indoor room}
		\vspace{3pt}
	\end{minipage}
	\begin{minipage}{0.15\textwidth}
	    \centering
		\includegraphics[width=\textwidth]  {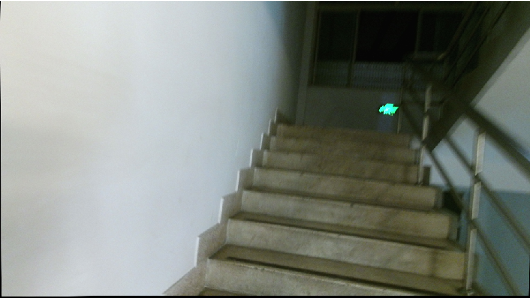}
		{\footnotesize (b) Indoor stairs}
		\vspace{3pt}
	\end{minipage}
	\begin{minipage}{0.15\textwidth}
	    \centering
		\includegraphics[width=\textwidth]  {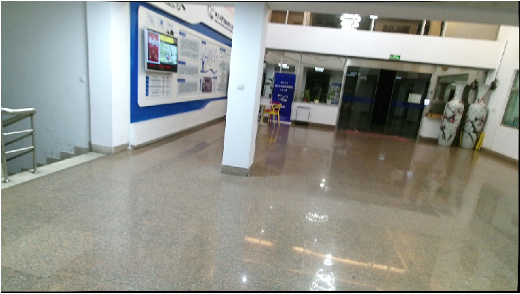}
		{\footnotesize (c) Indoor hall}
		\vspace{3pt}
	\end{minipage}
	\begin{minipage}{0.15\textwidth}
	    \centering
		\includegraphics[width=\textwidth]  {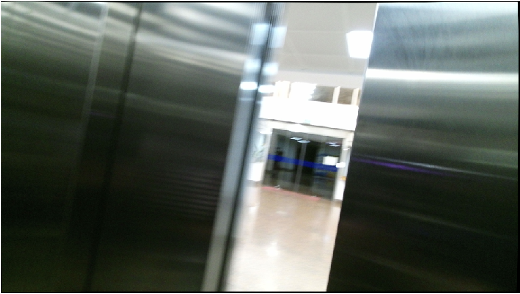}
		{\footnotesize (d) Indoor lift}
	\end{minipage}
	\begin{minipage}{0.15\textwidth}
	    \centering
		\includegraphics[width=\textwidth]  {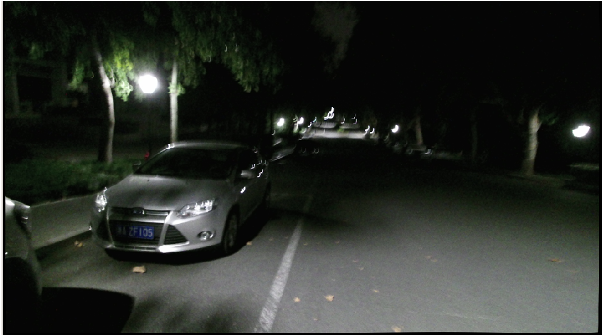}
		{\footnotesize (e) Outdoor road}
	\end{minipage}
	\begin{minipage}{0.15\textwidth}
	    \centering
		\includegraphics[width=\textwidth]  {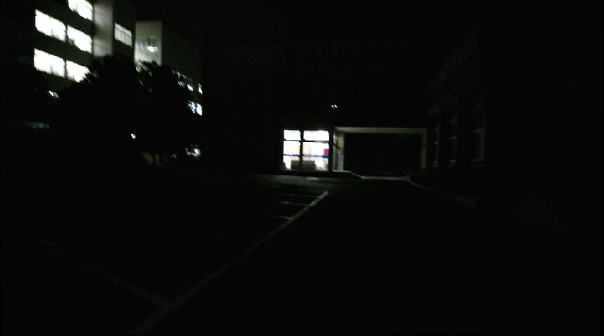}
		{\footnotesize (f) Outdoor building}
	\end{minipage}
	\caption{Data acquisition environments. }
	\label{fig:scenes}
	\vspace{-16pt}
\end{figure}

The dataset was acquired in indoor office environments and outdoors at night, as shown in Fig. \ref{fig:scenes}. 
The sequences are named according to these criteria: scene, illumination, motion, and the presence of a person in the image.

\subsubsection{Illumination}
\begin{figure}[t]
	\centering
	\includegraphics[width=0.49\textwidth]  {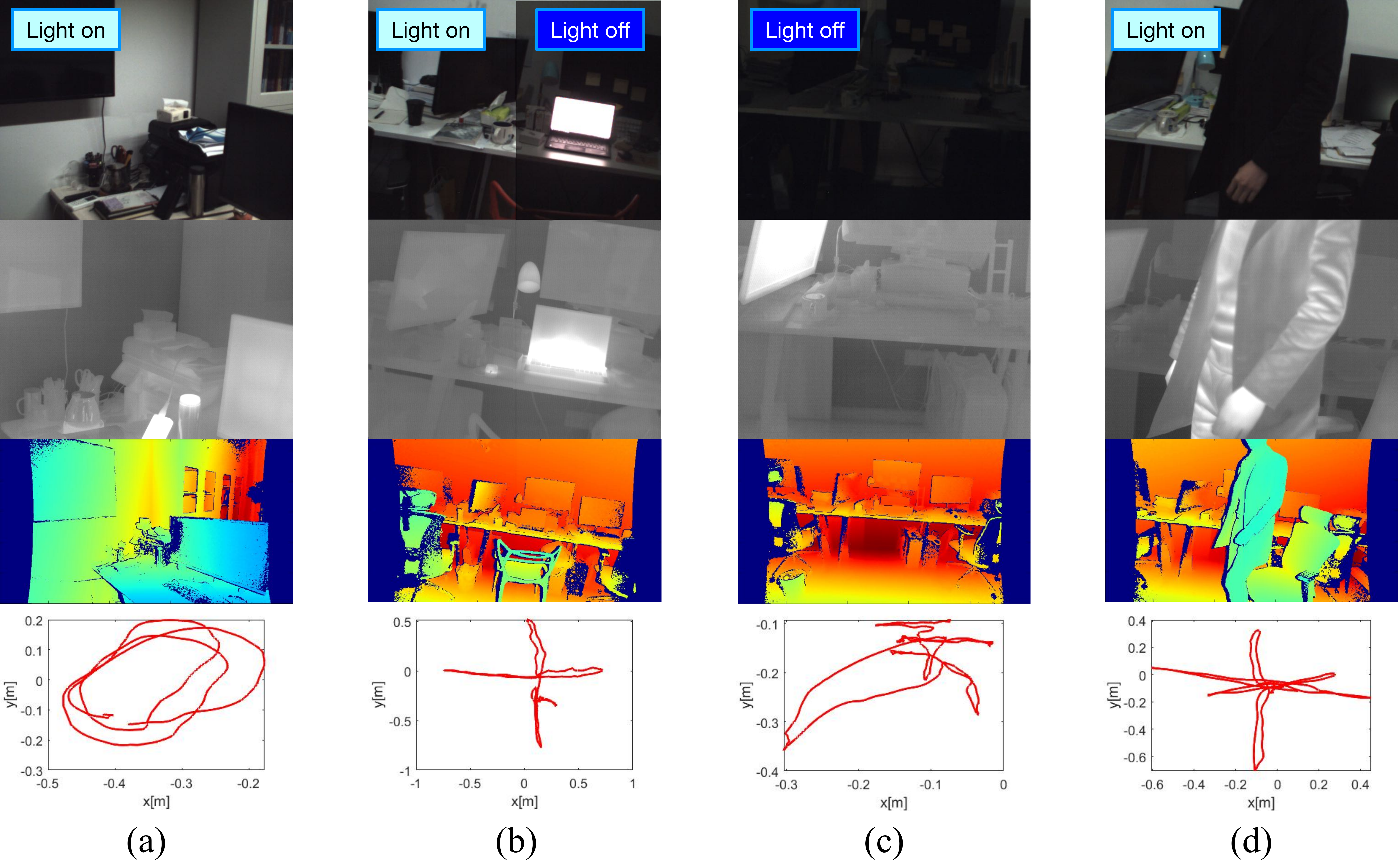}
	\caption{Example sequences. The images from the top to the bottom row are color images, thermal images, depth images, and ground-truth trajectories. The color of the depth images indicates the distance to the Kinect2. (a) desk2-circle. (b) desk1-halfsphere-ic. (c) desk1-rpy-dim2. (d) desk1-xyz-dy.}
	\label{fig:overview}
	\vspace{-16pt}
\end{figure}

Novel methods can be evaluated in environments with different illumination conditions as shown in Fig. \ref{fig:overview}.
In addition to the sequences recorded in bright environments, the sequences with the suffix \textsl{-ic} in the name contain data with lights turned on and off randomly. Moreover, for testing the performance of multi-spectral methods in extreme illumination conditions, the sequences with suffix \textsl{-dim} contain data in dark scenes. In the challenging sequences, the images are captured under complex illumination conditions.

\subsubsection{Motion}
These sequences with \textit{desk} are intended to evaluate methods with separated motions along and around the principal axes of the setup.
There are five types of camera motion: 
\begin{itemize}
\item[\textasteriskcentered] \textit{-halfsphere (hfsp)} denotes that the camera moves along the trajectory of a halfsphere with a diameter of 1m.
\item[\textasteriskcentered] \textit{-xyz} denotes that the camera moves approximately along the $x$, $y$, and $z$ axes.
\item[\textasteriskcentered] \textit{-rpy} denotes that the camera only rotates with roll, pitch, and yaw motion.
\item[\textasteriskcentered] \textit{-circle} denotes that the camera moves approximately around a circle.
\item[\textasteriskcentered] \textit{-static} denotes that the camera is nearly motionless.
\end{itemize}

\subsubsection{Person} Since the temperature of the objects in the office is almost uniform, the LWIR camera cannot provide rich textures. Since the temperature of the human body differs significantly from the environment, a person in the FoV of the camera can provide striking contrasts in textures in the thermal images.
Hence, in the sequences with the suffix \textit{-person}, there is a person sitting in front of a desk. Otherwise, the environment only contains static objects at room temperature.

\subsubsection{Dynamic} In the sequences with the suffix \textit{-dy}, a person is walking around in front of the setup to add disturbances. 
These data can be used to test the robustness of the multi-spectral system in dynamic environments since moving objects in the environment may lead to the failure of motion estimation.

\subsubsection{Scene}
The sequences recorded in the office room are labeled as \textit{desk1} and \textit{desk2}. \textit{magistrale} denotes sequences featuring a walk in a university building. The sequences with the suffix \textit{outdoor} are recorded around a university building. The suffix \textit{lift} indicates that the camera was moved to different floors via elevator.

More information about each sequence can be found on the website.

\subsection{Data format}

Each sequence is saved as a ROS bag file. 

\subsubsection{ROS Bag Files} raw data is recorded in the following topics:
\begin{itemize}
	\item /camera/image\_raw: color images from the standard camera
	\item /optris/thermal\_image: thermal images from the LWIR camera
	\item /camera/flag\_state: flag states of the LWIR camera
	\item /kinect2/qhd/image\_color\_rect: color images from the Kinect2
	\item /kinect2/qhd/image\_depth\_rect: depth images from the Kinect2
	\item /kinect2/sd/image\_ir\_rect: NIR images from the Kinect2
	\item /imu/data: IMU data
	\item /vrpn\_client\_node/RigidBody/pose: raw poses from the motion capture system
\end{itemize}
The meaning of most of these topics is self-explanatory. All data use the time in the ROS system as the timestamp. The /flag\_state of NUC is a type \textit{enum} and contains the flag signal of the LWIR camera with the following fields: \textit{FlagOpen, FlagClose, FlagOpening, FlagClosing}, and \textit{Error}. The last topic contains raw poses stored as both a vector and a quaternion from the motion capture system.

\section{EXPERIENCES}
\label{sec:exp}

This section presents an experimental evaluation with the proposed sequences to show how challenging this dataset can be. Since there is no open-source multi-spectral method, results were obtained from DSO (direct sparse odometry, direct method)~\cite{engel2017direct} and  ORB-SLAM3 (feature-based method)~\cite{campos2020orb}, which are two state-of-the-art monocular methods. The distinction between those two methods is that feature-based methods use the reprojection error of feature points. In contrast, direct methods exploit the photometric error of raw images directly. Besides, to show the benefit of LWIR, we evaluate both methods with a different spectrum.

As shown in Table \ref{tab:exp}, these two methods show different performance. Since ORB-SLAM3 only uses corner features, the sequences captured in the office without rich textures may not have enough uniform distribution features for ORB-SLAM3 for robust matching. Therefore, the ORB-SLAM3 cannot perform very well in these sequences with good illumination. It also failed in all sequences with varying and dim illumination, although it has a relocalization module. On the other hand, DSO can utilize the edge and the texture in dark environments, showing a better performance in environments with poor illumination. 

From the comparison between different spectrum 
we see that the method using LWIR shows better robustness but worse precision. This is because thermal images are independent from environment illumination but contain more noise from camera self-emission. For this reason, most uncooled LWIR cameras use NUC to eliminate the fixed noise, which may lead to fake tracking for the DSO method. Besides, the ORB feature is not designed for thermal images. Thus, the ORB feature descriptor works poorly.

\begin{table}[ht]
	
	\centering
	\caption{Comparison of the absolute trajectory error (ATE).
	}
	\label{tab:exp}
	{
	\begin{tabular}{lcc|cc}
			\toprule
			& \multicolumn{2}{c}{DSO} & \multicolumn{2}{c}{ORB-SLAM3} \\
			Sequence Name &  (RGB)  & {(LWIR)}  &  {(RGB)} &  {(LWIR)}  \\
			\midrule
	        \multicolumn{4}{l}{Bright illumination}       \\ \midrule
			desk1-halfsphere   & 0.1114 	& x     &  0.0186 & x \\
			desk1-rpy  & 0.0789   & x         &0.0929 & x  \\
			desk1-static  & x  & x & x & x  \\
			desk1-xyz  & 0.0086 & 0.2050 & 0.0165 & x  \\
			desk2-circle  & 0.0283  & x & 0.0214 & x  \\
			desk2-halfsphere  & 0.0095  & 0.3106  & 0.0191 & x  \\
			desk2-rpy  & 0.0747  & x & 0.0174 & x  \\
			desk2-static  & 0.0018  & x & 0.005193 & x  \\
			desk2-xyz  & 0.0065  & 0.005904 & 0.0098 & x  \\
			desk1-circle-person   &{0.0138}   & x      & 0.017505 & x  \\
			desk1-hfsp-person  & 0.0248   & 0.2909         & 0.0123 & x  \\
			desk1-rpy-person  & 0.1254  & x & 0.0336 & x  \\
			desk1-rpy-person-slow  & 0.0617  & x & 0.0550 & x \\
			desk1-static-person  & x & x & x & x \\
			desk1-xyz-person  & 0.0056  & 0.0280 & 0.0104 & 0.0484 \\
			desk1-halfsphere-dy  & 0.2553  & x & 0.1532 & x \\
			desk1-rpy-dy  & 0.0921   &  x &  x & x \\
			desk1-static-dy  &  x  & x  & x  & x \\
			desk1-xyz-dy  &  0.2677  & x  &  0.0177 & x \\
			\midrule
	        \multicolumn{4}{l}{Varying illumination}       \\ \midrule
			desk1-rpy-ic  & x   & x   & 0.0422  & x \\
			desk1-halfsphere-ic  & x   & 0.2472    & 0.2003  & x \\
			desk1-static-ic  & x  & x & x& x  \\
			desk1-static-ic-lampon  & x  & x & x& x  \\
			desk1-xyz-ic  & x  & 0.2858 & x& x  \\
			desk2-circle-ic  &  x  & x  &  x& x  \\
			desk2-rpy-ic  &  x  & x  &  x& x  \\
			desk2-static-ic  &  x  &  0.005904 & x  & x \\
			desk2-xyz-ic  & x  & 0.2343  &  x & x \\
			desk1-hfsp-ic-person   & x	  & 0.2558 &  x & x \\
			desk1-rpy-ic-person  & x  & 0.0851 &  x& x  \\
			desk1-static-ic-person  & x & x & x & x \\
			desk1-xyz-ic-person  & x  & x & x& x  \\
			\midrule
			\multicolumn{4}{l}{Dim illumination}       \\ \midrule
			desk1-halfsphere-dim  &  x  &  0.3792 &   x& x \\
			desk1-halfsphere-dim2  & 0.3200  & 0.3730  &  x& x  \\
			desk1-rpy-dim  &  x  & x  &  x & x \\
			desk1-rpy-dim2  &  x  &  x &  x& x  \\
			desk1-static-dim  &   x & x  & x & x  \\
			desk1-static-dim2  & x   & x  &  x & x \\
			desk1-xyz-dim  & 0.0130   &  0.2449 &  x & x \\ 
			desk1-xyz-dim2  &  0.2838  & 0.3674  &  x & x \\
			\midrule
			\multicolumn{4}{l}{Complex environments}       \\ \midrule
			magistale-*  &  x  &  x &   x& x \\
			outdoor-*  & x  & x  &  x& x  \\
			\bottomrule
	\end{tabular}}
	\vspace{-16pt}
\end{table}

\begin{figure}[t]
	\centering
	\begin{minipage}{0.23\textwidth}
	    \centering
		\includegraphics[width=\textwidth]  {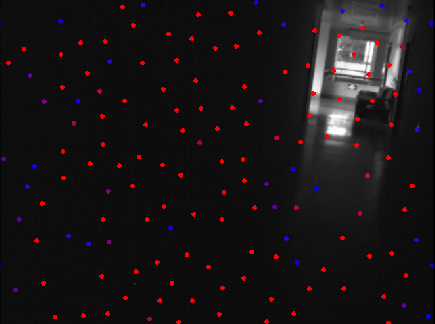}
		{\footnotesize Tracking on color images}
	\end{minipage}
	\begin{minipage}{0.23\textwidth}
	    \centering
		\includegraphics[width=\textwidth]  {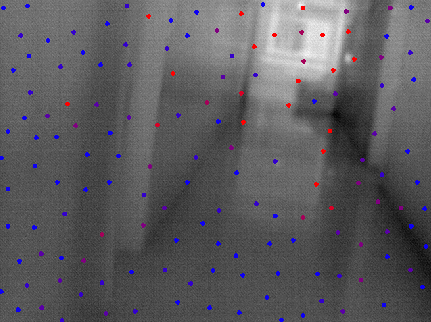}
		{\footnotesize Tracking on thermal images}
	\end{minipage}
	\caption{Tracking results from VINS-Mono. The red and blue points represent the reliable and unreliable points considered by VINS.}
	\label{fig:vins}
	\vspace{-16pt}
\end{figure}

For the sequences with \textit{magistrale} and \textit{outdoor}, both methods failed in tracking, because in these sequences there are scenes with different challenges, such as texture-less, large-scale, low illumination, and high dynamic range scenes. Since these sequences contain the IMU information, the VINS-Mono\cite{qin2018vins} method was also tested.
Similar to the results of DSO and ORB-SLAM3, VINS cannot complete the entire motion estimation either, as shown in Fig.~\ref{fig:vins}. However, the reasons for the failure of VINS-mono on color and thermal images are different. For color images, it is mainly due to the illumination, which can reduce the reliable features and led to the failure. For thermal images, although minor temperature variation can be captured, the LWIR camera still cannot provide high-quality texture information. As a result, it is difficult for VINS to obtain stable feature points on the thermal images to complete the estimation, even the initialization.

In summary, for this dataset the methods based on the standard camera cannot provide robust estimations when the environment's illumination becomes increasingly complex. In contrast, cameras that do not rely on ambient illumination can provide information about the environment to perform localization tasks. However, from the accuracy comparison, we can see that the unique mechanism of the LWIR camera impairs its performance in bright illumination compared with the standard camera. Meanwhile, the failure in all the magistrale and outdoor environments suggests that the only way to achieve robust localization in complex environments is to combine several different sensors. Hence this dataset's challenge increases in the order of bright, dark(desk-dim), varying(desk-ic), and complex(magistrale, outdoor) environments.

\section{CONCLUSIONS}
\label{sec:conclusion}

In this paper, a novel dataset was proposed for evaluating the performance of multi-spectral motion estimation methods. The dataset includes sequences captured from different scenes with a diverse set of illumination conditions. In the set-up, the multi-spectral cameras were in hardware-synchronization and well-calibrated. Highly accurate ground-truth poses were provided by a motion capture system for evaluation. Besides, depth images were provided for studying cross-modality stereo matching. Some commonly used robustness metrics were also proposed. This dataset is publicly available with both raw and calibration data. 
It is hoped that this dataset can facilitate the development of multi-spectral motion estimation.

Moreover, during the preparation of the dataset, some experience was gained. 1. United acquisition program. As far as possible, the drivers of all sensors should be integrated into one program instead of distributed in separate software. Through a unified acquisition procedure, the acquisition efficiency can be improved, while the unpredictable problems that occur during the acquisition process can be reduced. 2. Data check program. After the acquisition, the check program can quickly find whether there are acquisition problems, such as data loss, to improve the acquisition efficiency. We hope that our experience will be helpful to everyone.





\balance

{\small
\bibliographystyle{IEEEtran}
\bibliography{ref}
}

\end{document}